\documentclass[10pt,twocolumn,letterpaper]{article}

\usepackage[submission]{cvpr} %
\usepackage{times}
\usepackage{epsfig}
\usepackage{graphicx}
\usepackage{amsmath}
\usepackage{amssymb,amsfonts}

\usepackage{xspace}
\usepackage[misc]{ifsym}
\usepackage{multicol}
\usepackage{stfloats}
\usepackage{makecell}
\usepackage{pifont}
\usepackage{colortbl}
\usepackage{tabularx}
\usepackage{caption}

\usepackage{listings}
\usepackage{mathtools}
\usepackage{cite}

\usepackage{algorithmic}
\usepackage{algorithm}
\usepackage{array}
\usepackage[caption=false]{subfig}
\usepackage{textcomp}
\usepackage{url}
\usepackage{verbatim}
\usepackage{graphicx}
\usepackage{color}
\usepackage{booktabs}
\usepackage{multirow}
\usepackage{bbding}

\newcolumntype{x}[1]{>{\centering\arraybackslash}p{#1pt}}

\newlength\savewidth
\newcommand{\tablestyle}[2]{\setlength{\tabcolsep}{#1}\renewcommand{\arraystretch}{#2}\centering\footnotesize}

\usepackage[pagebackref=true,breaklinks=true,letterpaper=true,colorlinks,bookmarks=false]{hyperref}

\def\paperID{6032} 
\def\confName{CVPR}
\def\confYear{2024}
\newcommand\blfootnote[1]{
    \begingroup
    \renewcommand\thefootnote{}\footnote{#1}
    \addtocounter{footnote}{-1}
    \endgroup
}

\title{Adapting Short-Term Transformers for Action Detection in Untrimmed Videos}

\author{Min Yang\textsuperscript{1} \quad \quad Huan Gao\textsuperscript{2} \quad \quad Ping Guo\textsuperscript{3} \quad \quad Limin Wang\textsuperscript{1,4,~\Letter}\\
$^1$State Key Laboratory for Novel Software Technology, Nanjing University  \\
\quad $^2$Inchitech \quad $^3$Intel Labs China \quad $^4$Shanghai AI Lab \\
{\tt\small  yangminmcg1011@hotmail.com, gaohuan@inchitech.com, ping.guo@intel.com, lmwang@nju.edu.cn} \\
\textbf{\normalsize\url{https://github.com/MCG-NJU/ViT-TAD}}
}

\begin{document}
\maketitle

\begin{abstract}
Vision Transformer (ViT) has shown high potential in video recognition, owing to its flexible design, adaptable self-attention mechanisms, and the efficacy of masked pre-training. 
Yet, it remains unclear how to adapt these pre-trained short-term ViTs for temporal action detection (TAD) in untrimmed videos. The existing works treat them as off-the-shelf feature extractors for each short-trimmed snippet without capturing the fine-grained relation among different snippets in a broader temporal context.
To mitigate this issue, this paper focuses on designing a new mechanism for adapting these pre-trained ViT models as a unified long-form video transformer to fully unleash its modeling power in capturing inter-snippet relation, while still keeping low computation overhead and memory consumption for efficient TAD. 
To this end, we design effective cross-snippet propagation modules to gradually exchange short-term video information among different snippets from two levels. 
For inner-backbone information propagation, we introduce a cross-snippet propagation strategy to enable multi-snippet temporal feature interaction inside the backbone.
For post-backbone information propagation, we propose temporal transformer layers for further clip-level modeling.
With the plain ViT-B pre-trained with VideoMAE, our end-to-end temporal action detector (ViT-TAD) yields a very competitive performance to previous temporal action detectors, riching up to 69.5 average mAP on THUMOS14, 37.40 average mAP on ActivityNet-1.3 and 17.20 average mAP on FineAction.
\vspace{-6mm}
\end{abstract}

\blfootnote{\Letter: Corresponding author.}

\section{Introduction}

As an important task in video understanding, temporal action detection (TAD)~\cite{thumos14,anet,fineaction} aims to localize all action instances and recognize their categories in a long untrimmed video. Most TAD methods~\cite{basictad,bmn,ActionFormer,afsd,stpt,re2tal} rely on the pre-trained action recognition networks (backbones) to extract short-term features for each snippet and then apply the TAD heads on top of feature sequence for action detection in long-form videos. In this pipeline, the feature extracted from the backbone is of great importance in the final TAD performance. To obtain powerful features, the existing TAD methods~\cite{ActionFormer,basictad,stpt,re2tal} have tried different kinds of backbones, from CNN-based backbones~\cite{tsn,i3d} to Transformer-based ones~\cite{videoswinT,mvit}. Recently, the transformer~\cite{attention} has become a promising alternative to CNN in modern TAD pipeline design, thanks to its effective self-attention operations. Furthermore, the flexible vision transformer (ViT)~\cite{vit} enables self-supervised masked pre-training on extensive video datasets~\cite{videomae}, resulting in a more robust video representation learner.

\begin{figure}[!t]
  \includegraphics[width=0.5\textwidth]{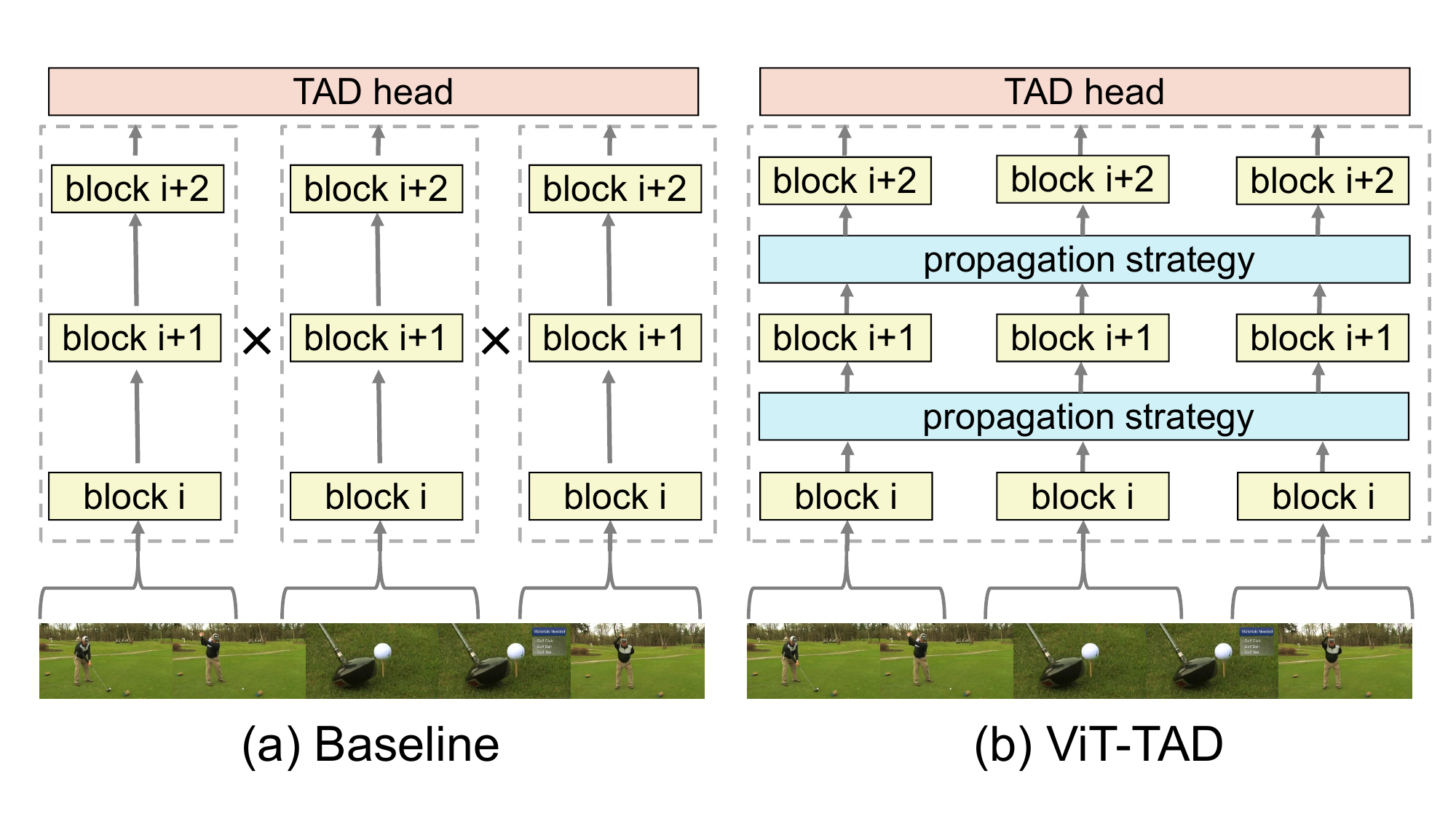}
  \caption{
\textbf{Different input processing between baseline and our ViT-TAD.} 
The dashed box illustrates the feature modeling within the backbone.
In contrast to the baseline approach which models each snippet individually, our ViT-TAD allows snippets to collaboratively interact with each other during the modeling process within the backbone.
}
\label{fig:intro}
\vspace{-4mm}
\end{figure}

Several TAD methods~\cite{stpt,TALLFormer,re2tal,faster-tad} have tried to apply Transformer-based backbones, such as VideoSwin~\cite{videoswinT} and MViT~\cite{mvit}, to the task of TAD. Unlike CNN, these transformer backbones~\cite{videoswinT,mvit,vit} present challenges when applied directly to untrimmed video modeling due to the quadratic computational cost associated with self-attention operations~\cite{attention}. Therefore, the existing methods often treat the transformer network as the {\bf off-the-shelf} feature extractor for each snippet independently, as shown in Fig~\ref{fig:intro}(a), thereby neglecting the intricate temporal relationships across snippets and failing to fully harness the end-to-end representation learning power of Transformers for TAD. Meanwhile, 
the hierarchical transformer architecture of VideoSwin or MViT they choose which is hard to benefit from the powerful masked video pre-training~\cite{videomae} on large-scale unlabeled videos. Therefore, it still remains unclear {\em how to effectively adapt the short-term {\bf plain ViT} to untrimmed video action detection with the ability of capturing {\bf cross-snippet temporal structure}} in an end-to-end manner.

To this end, we focus on building a simple and general temporal action detector based on the plain ViT backbone~\cite{vit}. Rather than employing the pre-trained short-term ViTs as the snippet feature extractors, we design an efficient mechanism of adapting them to model longer video consisting of multiple snippets within a unified transformer backbone, as shown in Fig~\ref{fig:intro}(b). Following the success of ViT in object detection~\cite{vitdet}, we first divide video into non-overlapping snippets and apply both intra-snippet and inter-snippet self-attention operations to keep a trade-off between representation power and computational cost. In this way, our unified transformer backbone can model multiple snippets as a whole and capture the fine-grained and holistic temporal structure for the TAD task.

Specifically, we apply dense self-attention on all tokens within each snippet to capture their spatiotemporal relations in intra-snippet blocks and propose a cross-snippet propagation module to aggregate global temporal information in a position-wise manner in inter-snippet blocks. These two kinds of blocks are stacked alternatively to gradually exchange temporal information within long videos. In addition, inspired by the design of DETR~\cite{detr}, we devise the post-backbone information propagation module after our Transformer-based backbone, which is composed of several temporal transformer layers to aggregate global temporal information. This post-backbone propagation module can effectively enlarge the temporal receptive field and capture global context. Equipped with simple TAD head such as BasicTAD~\cite{basictad} and AFSD~\cite{afsd}, our final temporal action detector, termed as {\bf ViT-TAD}, enjoys a simple yet effective design with end-to-end training. In particular, our ViT-TAD can embrace the powerful self-supervised masked pre-training~\cite{videomae} and yield state-of-the-art performance on the challenging datasets THUMOS14 and ActivityNet-1.3. In summary, our contributions are as follows:
\begin{itemize}
    \item
    We introduce ViT-TAD,  the first end-to-end TAD framework that utilizes the plain ViT backbone. 
    Through the incorporation of a straightforward inner-backbone information propagation module, ViT-TAD can effectively treat multiple video snippets as a unified entity, facilitating the exchange of temporal global information. 
    \item With a simple TAD head and careful implementation, we can train our ViT-TAD in an end-to-end manner under the limited GPU memory. This simple design fully unleashes the modeling power of the transformer and embraces the strong pre-training of VideoMAE~\cite{videomae}.
    \item The extensive experiments on THUMOS14~\cite{thumos14}, ActivityNet-1.3~\cite{anet} and FineAction~\cite{fineaction} demonstrate that our simple ViT-TAD outperforms the previous state-of-the-art end-to-end TAD methods.
\end{itemize}

\section{Related Work}
\paragraph{\bf{Transformer in Action Recognition.}}  
Action recognition is an important task in video understanding. 
With the success of the self-attention mechanism in computer vision, several works tried to apply the transformer to their structures. 
Specifically, VTN~\cite{vtn} and STAM~\cite{stam} introduced temporal transformers to encode frame-level relationships between features. ViViT~\cite{vivit} and TimeSformer~\cite{timesformer} factorized along spatial and temporal dimensions on the granularity of the encoder. SMAVT~\cite{smavt} aggregated information from tokens at the same spatial location within a local temporal window. Some works~\cite{mvit,videoswinT} tried to re-introduce hierarchical designs into transformer inspired by ConvNet. Among them, MViT~\cite{mvit} presented a hierarchical transformer structure to progressively shrink the spatiotemporal resolution of feature maps and increase channels as the network structure goes deeper. VideoSwin~\cite{videoswinT} used shifted window attention to enable information propagation inspired by Swin Transformer~\cite{swinT}. 
Furthermore, with the development of self-supervised learning~\cite{videomae}, Transformer-based methods benefit from larger training data and achieve better results than CNN-based methods on the action recognition task.

\vspace{-2mm}
\paragraph{\bf{Transformer in Temporal Action Detection.}}
With the development of transformer, more and more TAD approaches began to apply it or its variants into TAD head~\cite{e2e-tadtr,rtd,ActionFormer,react} or backbone~\cite{stpt,TALLFormer,re2tal,faster-tad}. Different from increasingly successful applications of transformer in TAD head, few works seriously explored the application of transformer in the backbone. Existing works~\cite{faster-tad,TALLFormer,re2tal} treated Transformer-based backbones as ``black box'' and applied them as short-term feature extractors.  
STPT~\cite{stpt} attempted to explore the inner modeling of a hierarchical Transformer-based backbone by inserting new blocks in it. However, it needs additional pre-training on the action recognition dataset
which fails to embrace the benefits of pre-trained big models~\cite{internvideo,tubevit}. 
In this work, we try to design a new mechanism for adapting a pre-trained short-term snippet-level non-hierarchical ViT-based backbone as a unified clip-level video transformer.

\begin{figure*}[!t]
  \includegraphics[width=1.0\textwidth]{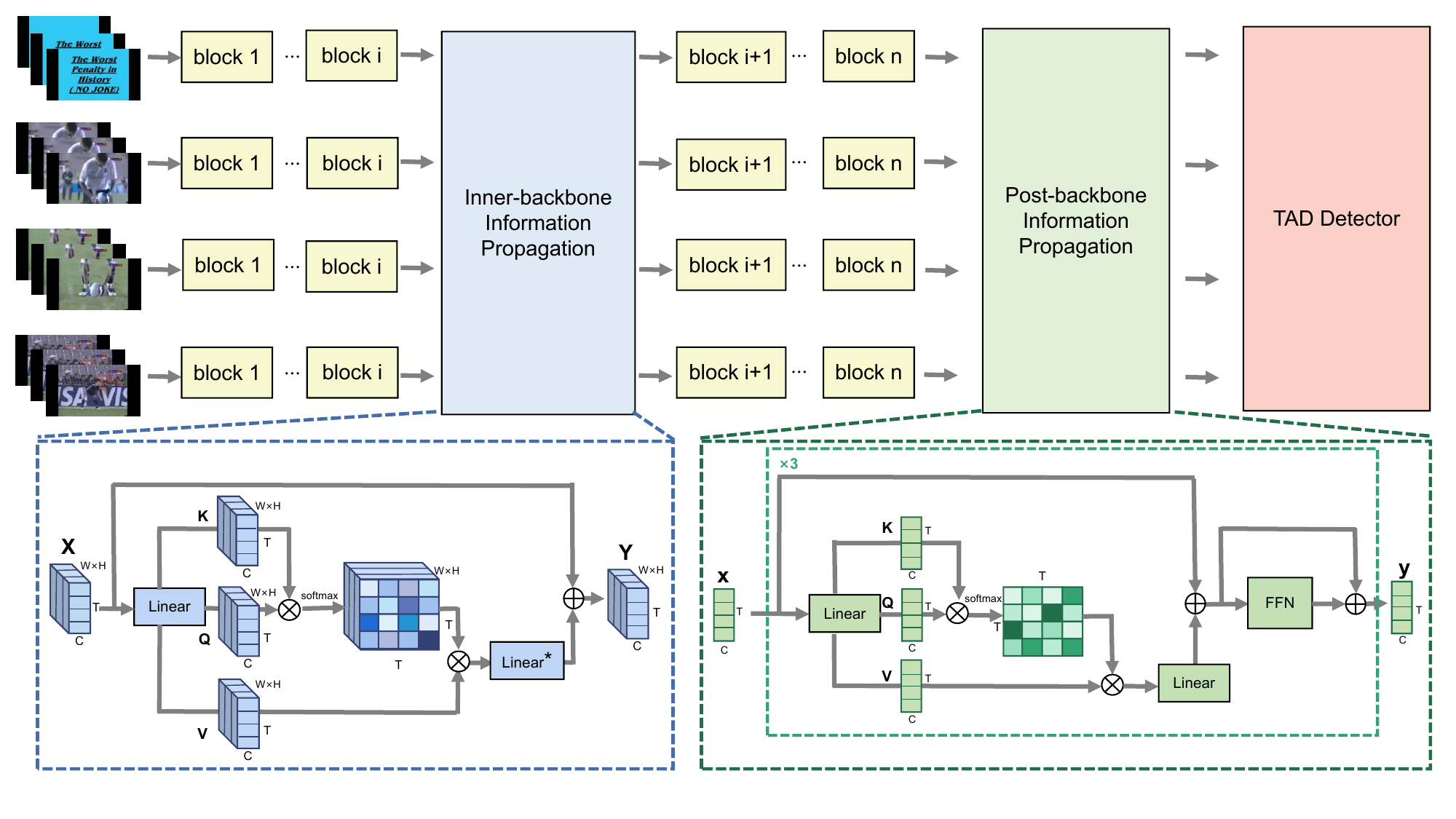}
  \caption{
\textbf{Overview of ViT-TAD.}
We suppose the ViT-based backbone has $n$ blocks and divide them into several subsets. Each subset has $i$ blocks. 
We divide a video clip into several snippets and send each into the backbone for feature extraction. 
We perform temporal feature interaction among all snippets through the inner-backbone information propagation strategy. We further conduct clip-level modeling to refine clip-level features through the post-backbone information propagation strategy. 
$*$ means the last layer is initialized as zero.
}
\label{fig:ViT-TAD}
\vspace{-5mm}
\end{figure*}

\vspace{-2mm}
\paragraph{\bf{End-to-End Temporal Action Detection.}}
Due to GPU memory constraints, most existing methods treated TAD~\cite{bmn,bsn,gtad,vsgn} as a two-step approach separating feature modeling and action detection. Since temporal features are extracted using backbones that are pre-trained on the action recognition dataset, there is a gap between these two different tasks, leading to sub-optimal performance. In order to address this gap, incorporating the backbone into the network's training process forms an end-to-end TAD framework~\cite{afsd,TALLFormer,pbrnet,stpt,re2tal}. Simultaneously updating the network weights of both parts in the two-step approach will easily exceed the GPU memory limit. 
To address this issue, some methods~\cite{afsd,pbrnet,stpt} tried to downscale the frame resolution, while others tried to keep the resolution of the original video frame through improvements in training mechanisms. Among them, TALLFormer~\cite{TALLFormer} attempted to build an offline memory bank to store features that cannot be updated synchronously and only update a portion of features during end-to-end training. 
Re$^{2}$TAL~\cite{re2tal} built a backbone with reversible modules to save memory consumption. The concurrent work AdaTAD~\cite{1000frames} achieved the purpose of saving video memory by only training the adapter and freezing the backbone.
Due to the existence of positional encoding, how to fine-tune the TAD model when spatiotemporal resolution differs from the pre-trained Transformer-based model is a problem. TALLFormer~\cite{TALLFormer} and Re$^{2}$TAL~\cite{re2tal} circumvented this problem by keeping the same spatiotemporal resolution as the pre-trained backbone, while STPT~\cite{stpt} pre-trained its own backbone. We adopt the downscaling frame resolution strategy and use the operation from DeiT~\cite{deit} to fix the positional encoding across resolutions to better use the pre-trained weight of the ViT-based backbone.

\section{ViT-TAD} 

\noindent {\bf Overview.} We describe the details of our ViT-TAD pipeline shown in Fig~\ref{fig:ViT-TAD}.
Formally, for each input video $V \in R^{T \times H \times W \times 3}$, where $T$, $H$ and $W$ represent the number, height and width of RGB frames respectively. 
We divide it into $N_s$ non-overlapping snippets $s=\{s_j\}^{N_s}_{j=1}$, where $s_j \in R^{(T / N_s) \times H \times W \times 3}$. Then we feed each non-overlapping snippet $s_j$ into consecutive blocks of the ViT-based backbone. In the baseline approach (shown in Fig~\ref{fig:intro}(a)) which is adopted by the existing methods~\cite{TALLFormer,re2tal}, each snippet has no interaction with each other during fine-tuning, so we propose \emph{inner-backbone information propagation} module to enable cross-snippet interaction between all snippets. 
Given that the inner-snippet and cross-snippet modeling process within the backbone is gradual and sparse, resulting in relatively small receptive fields, it becomes necessary to enhance the final snippet-level features by effectively expanding their temporal receptive field.
So we propose the \emph{post-backbone information propagation} module composed of several temporal transformer layers to refine snippet-level temporal features with the global context.
Finally, the enhanced features will be sent to the TAD head for action detection, producing $K$ predicted actions $\{(s_i,e_i,c_i,p_i)\}_{i=1}^{K}$ as final detection results, where $s_i,e_i,c_i,p_i$ represent the start timestamp, end timestamp, category and confidence score of $i^{th}$ action instance. 

\subsection{Inner-backbone Information Propagation}
We design a cross-snippet propagation module for inner-backbone information exchange. Unlike hierarchical backbone structures like VideoSwin~\cite{videoswinT}, we do not shift the temporal windows across layers. We use very few blocks that can go across snippets to allow information propagation. Inspired by ViT-Det~\cite{vitdet}, we evenly split a pre-trained ViT-based backbone into several subsets, and then insert a cross-snippet propagation module after the last block of each subset to enable information propagation. For the simplicity of the method, 
we adopt two simple but useful blocks as our cross-snippet propagation modules, coined as \emph{Local Propagation Block} and \emph{Global Propagation Block}. 

\vspace{-5.0mm}
\paragraph{\bf{Local Propagation Block.}}
We choose the bottleneck architecture, comprising three 3D convolutions and an identity shortcut, as our local block. 
The last normalization layer within this block is initialized to zero, establishing an initial identity state for the block. This design permits integration into any place without breaking the pre-trained model. 
Although the cross-snippet modeling capacity of convolutions is limited, it can still exchange cross-snippet information gradually via the stacking of multiple such modules. 

\begin{figure}[!t]
  \includegraphics[width=0.45\textwidth]{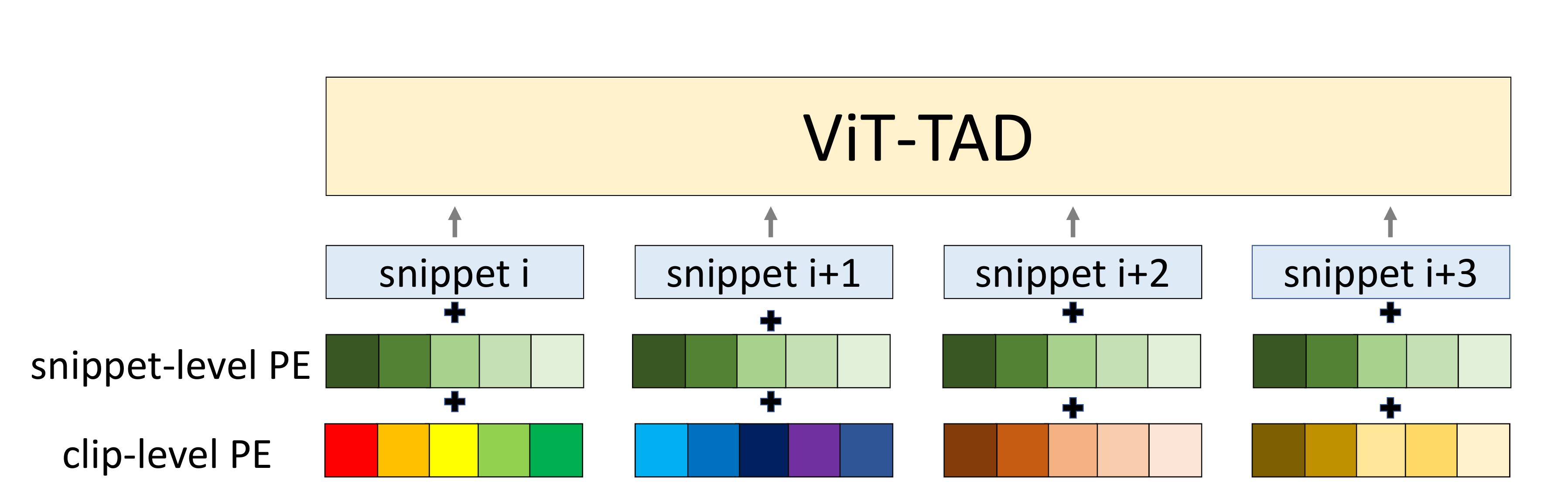}
  \caption{
\textbf{Temporal positional encoding for all snippets.} Original input consists of a snippet and its snippet-level PE. We add a learnable additional temporal PE, called clip-level PE here. PE is short for positional encoding.
}
\label{fig:pe}
\vspace{-3mm}
\end{figure}

\vspace{-5.0mm}
\paragraph{\bf{Global Propagation Block.}}
Self-attention mechanism is naturally suitable for long-term modeling, so we choose it to build a global block shown in Fig~\ref{fig:ViT-TAD}. To keep low computation overhead and memory consumption, we perform temporal self-attention among all video snippets after the last block of each stage. It is noteworthy that spatial dimensions do not participate in self-attention where width($W$) and height($H$) are directly flattened to batch dimension. 
Specifically, we obtain the sequence of clip-level spatiotemporal features $X \in R^{(W \times H) \times T \times C}$ and split them into $W \times H$ temporal features $x \in R^{T \times C}$, where $X=\mathrm{concat}(x_1,x_2,...,x_{W \times H})$ and each $x$ represents the clip-level temporal feature of a certain spatial location. For $x$ in each spatial location, we formulate our self-attention mechanism as:
\begin{equation}
\begin{split}
    q = xw_q, k = xw_k, v= xw_v, \\
    y = \mathrm{softmax}(qk^T/\sqrt{C_q})v,
\end{split}
\end{equation}
where $w_q \in R^{C \times C_q}$, $w_k \in R^{C \times C_k}$, $w_v \in R^{C \times C_v}$, $C_q = C_k = C/m$ and $m$ is the number of attention head. 
Our ViT-TAD complexity is reduced from $O(H^2W^2L_s^2N_sC)$ to $O(HWT^2C)$ where $L_s = T / N_s$. 
Finally, we obtain $Y = \mathrm{concat}(y_1,y_2,...,y_{W \times H}) \in R^{(W \times H) \times T \times C}$. 
After the application of multi-head dot-product attention, a linear layer is required to attend to information from distinct representation sub-spaces and positions jointly. Similar to the local propagation block, the last linear layer in the global block is initialized as zero and an identity shortcut is employed. 
Since each spliced short-term snippet in the clip shares the same temporal positional encoding, concatenating snippet-level positional encoding cannot capture the long-term temporal order of the time series. Therefore, we propose additional clip-level positional encoding to depict the temporal sequence's overall order within the video. As shown in Fig~\ref{fig:pe}, we add a learnable clip-level positional encoding to each temporal position of the snippet.

\subsection{Post-backbone Information Propagation} 
Given that modeling snippet-level features in the backbone with an inner-backbone information propagation strategy is a sparse and gradual process. 
we have designed a post-backbone information propagation module to ensure adequate feature interaction among snippets and incorporate clip-level modeling. 
Inspired by the encoder designed by DETR~\cite{detr}, we build $L$ temporal transformer encoder layers to allow snippets to interact with each other, as
shown in Fig~\ref{fig:ViT-TAD}. It is worth noting that the spatial dimension of each snippet feature has been squeezed with average pooling before this propagation module. 
Formally, given the video features $F$ concatenated by several snippets of feature provided by backbone, several temporal transformer encoder layers are adopted to enhance these features: $F^{i} = \mathrm{encoder}(F^{i-1})$, where $i \in [1,L]$ and L is the number of encoder layers, $F^{0}=F$ and $L=3$ in our experiment. 

\subsection{TAD Head}
Our proposed ViT-TAD framework is a simple and general temporal action detection pipeline. In principle, it is compatible with any TAD head for converting features into predictions. Following the previous practice in TAD, we choose a one-stage anchor-free method from BasicTAD~\cite{basictad} on THUMOS14 and FineAction, a two-stage anchor-free method from AFSD~\cite{afsd} on ActivityNet-1.3. 
 
\noindent {\bf Discussion with ViT-Det.} In spirit, our work is similar to ViT-Det~\cite{vitdet} which also has similar findings in the field of object detection where restricted window attention is not enough to capture object information in a global view. 
However, dynamic motion scenes are different from static object scenes, so our ViT-TAD behaves differently in several ways. {\em First}, temporal continuity makes action content highly correlated with long-range context, making the global block perform better than the local block which is different from ViT-Det. 
{\em Second}, we design post-backbone propagation to enhance further temporal context modeling based on the first finding, which is missing in ViT-Det. {\em Finally}, unlike spatial downsampling in ViT-Det, our ViT-TAD exhibits the same temporal downsampling rate and pattern as the previous CNN-based TAD method (e.g., BasicTAD~\cite{basictad}). This is the reason why we directly use the same FPN with BasicTAD without specific exploration, but ViT-Det needs further study. {\em In addition}, the TAD community often tend to avoid designing end-to-end detection pipeline, but instead devise complicated temporal modules on outdated pre-extracted features (e.g., I3D of 6 years ago). Building a neat end-to-end TAD pipeline and embracing more powerful models is more urgent and we make a meaningful attempt to facilitate TAD research in this direction.

\section{Experiments}
\paragraph{\bf{Datasets and Evaluation Metric.}}
We perform extensive experiments on THUMOS14~\cite{thumos14}, ActivityNet-1.3~\cite{anet} and FineAction~\cite{fineaction} to demonstrate the effectiveness of ViT-TAD. 
\textbf{THUMOS14} is a commonly-used dataset in TAD, containing 200 validation videos and 213 test videos with labeled temporal annotations from 20 categories. 
\textbf{ActivityNet-1.3} is a large-scale dataset containing 10,024 training videos and 4,926 validation videos belonging to 200 activities. \textbf{FineAction} is a newly collected large-scale fine-grained TAD dataset containing 57,752 training instances from 8,440 videos and 24,236 validation instances from 4,174 videos and 21,336 testing instances from 4,118 videos.
Following previous work, we report the \textbf{mean average precision (mAP)} with tIoU thresholds [0.3:0.1:0.7] for THUMOS14 and [0.5:0.05:0.95] for ActivityNet-1.3 and FineAction.  
\text{Avg} is average mAP on these thresholds.

\vspace{-5.0mm}
\paragraph{\bf{Implementation Details.}}

For THUMOS14 and FineAction, we sample each video clip with temporal windows of 32 seconds covering 99.7\% of action instances for THUMOS14 and 48 seconds covering 97.0\% of action instances for FineAction. 
We sample each video clip with 256 frames for THUMOS14 and 384 frames for FineAction at 8 FPS and then divide each clip into 16 and 24 snippets each, while each snippet consists of 16 frames. Due to the limited GPU memory, we resize frame's original size into short-128 (the short side of the frame is set to 128) and set the crop size (the size of the cropped images) to $112 \times 112$ for THUMOS14 in ablation experiments and report final results in short-180 original size and $160 \times 160$ cropping size. Inspired by DeiT~\cite{deit}, we need to adapt the positional embeddings to smaller spatial resolution with bicubic interpolation. For TAD head, we choose a one-stage anchor-free method from BasicTAD~\cite{basictad} for its astonishing detection performance. We train the model using SGD with a momentum of 0.9 and weight decay of 0.0001 on 8 TITAN Xp GPUs.
The batch size is set to 2 for each GPU. 
For ActivityNet-1.3, we resize all videos to 768 frames and then treat these frames as a video clip. Similarly, we divide each clip into 48 snippets and each snippet consists of 16 frames.
For TAD head, we choose a two-stage anchor-free method from AFSD~\cite{afsd} for its more accurate boundary regression. 
We resize the frame's original size into short-180 and set the crop size to $160 \times 160$. We adopt positional embeddings to smaller spatial resolution with bicubic interpolation. We train the model using AdamW~\cite{adamw} with a learning rate of 0.0002 and weight decay of 0.0001 on 8 TITAN Xp GPUs. The batch size is set to 1 for each GPU.

\begin{table}[]
\resizebox{0.47\textwidth}{!}{
\begin{tabular}{cc|ccccc|c}
\hline
prop. strat. & blk. num & 0.3 &0.4& 0.5  & 0.6&0.7 & Avg \\ \hline
none        &   -    & 74.6 & 70.2 & 62.6 & 51.3  & 38.4  & 59.4        \\ \hline
\multirow{2}{*}{global} & 4  & 78.4 & 74.5 &  66.6 & 53.4 & 38.7  &  \textbf{62.3}       \\
  & 12  & 76.5 & 71.6 & 63.3  & 51.9 &  38.4 &  60.3       \\ \hline
\multirow{2}{*}{local} & 4  & 75.9 & 71.6 & 64.7 & 53.3& 38.3 &  60.8      \\ 
 & 12  & 77.1 & 72.8 & 65.4 & 54.2& 40.6 &  \textbf{62.0}   \\ \hline
\end{tabular}
}
 \vspace{2mm}
\caption{\textbf{Study on inner-backbone information propagation}. Both propagation strategies and different numbers of blocks are explored. ``none'' is our baseline without propagation. 
}
\label{table:different-num-block}
\end{table}

\begin{table*}[t] \centering \vspace{-2.5em}
        \subfloat[\textrm{Effect of clip-level temporal positional encoding}\label{tbl:positional-encoding}]{
		\tablestyle{2pt}{1.05}
            \setlength{\tabcolsep}{0.5mm}
            \centering
            \begin{tabular}{c|ccc|c}
            \hline
            temp. pos. enc. & 0.3  & 0.5  & 0.7 & Avg \\ \hline
            \checkmark  & 78.4 & 66.6  & 38.7 &  \textbf{62.3}      \\
                  & 77.9  & 64.4  & 38.8 &  61.3      \\ \hline
            \end{tabular}
            }
        \hspace{2mm}
        \vspace{-1mm}
         \subfloat[\textrm{Comparison between 1D strategy and 3D strategy. Both strategies are shown in Fig~\ref{fig:1d_3d}.}\label{tbl:attn3d}]{
        \tablestyle{2pt}{1.05}
        \centering
        \begin{tabular}{c|ccc|cc}
        \hline
       prop. strategy & 0.3 & 0.5  & 0.7 & Avg & Mem. \\ \hline
        none               & 74.6 & 62.6   & 38.4    & 59.4   & 8.3GB    \\ \hline
        1D & 78.4 & 66.6 & 38.7 &  \textbf{62.3} &  9.3GB  \\
        3D & 78.0 & 65.5 & 39.7 &  62.2 & 27GB\\ \hline
        \end{tabular}
        }
        \vspace{2mm}
        \subfloat[\textrm{Study on the locations of global blocks. Here we choose 4 global blocks strategy.}\label{tbl:locations-of-propagation-blocks}]{
	\tablestyle{2pt}{1.05}
        \centering
        \begin{tabular}{c|ccc|c}
        \hline
        propagation location & 0.3 & 0.5 & 0.7 & Avg \\ \hline
        evenly 4 blocks & 78.4 & 66.6 & 38.7 &  \textbf{62.3}  \\
        first 4 blocks  &  75.2     &  64.1   &   38.5   & 60.1\\
        last 4 blocks   &   77.6    &  66.0    &  39.6    & 61.8\\ \hline
        \end{tabular}
        }
		\caption{\small \textbf{Study on global propagation strategy}. ``none'' is our baseline without propagation. 
         } 
		\label{tab:ablations}
\end{table*}

\subsection{Ablation Study}
 In this section, we perform ablation experiments for ViT-TAD on THUMOS14. Both ViT-S and ViT-B we adopted are pre-trained on Kinetics-400~\cite{k400} provided by~\cite{videomae_v2}. 
\vspace{-5mm}
\paragraph{\bf{Study on Inner-backbone Information Propagation.}}
Table~\ref{table:different-num-block} ablates our inner-backbone information propagation strategy. These propagation blocks are placed in the backbone evenly and kernel size is set to $(3 \times 3 \times 3)$ for the local propagation block. We compare our global and local propagation blocks with the baseline, and both of them perform better than the baseline in 4-block settings. We further explore the number of propagation blocks and find that the local strategy benefits from having more blocks, but the global strategy does the opposite. There are two plausible explanations for this. 
First, the model's ability to learn valuable information is hampered by frequent global interactions. Second, the self-attention operation cannot use prior information like the convolutional operator. To enhance temporal interaction, a more in-depth exploration is warranted.

\begin{figure}[!t]
 \vspace{-5mm}
  \includegraphics[width=0.45\textwidth]{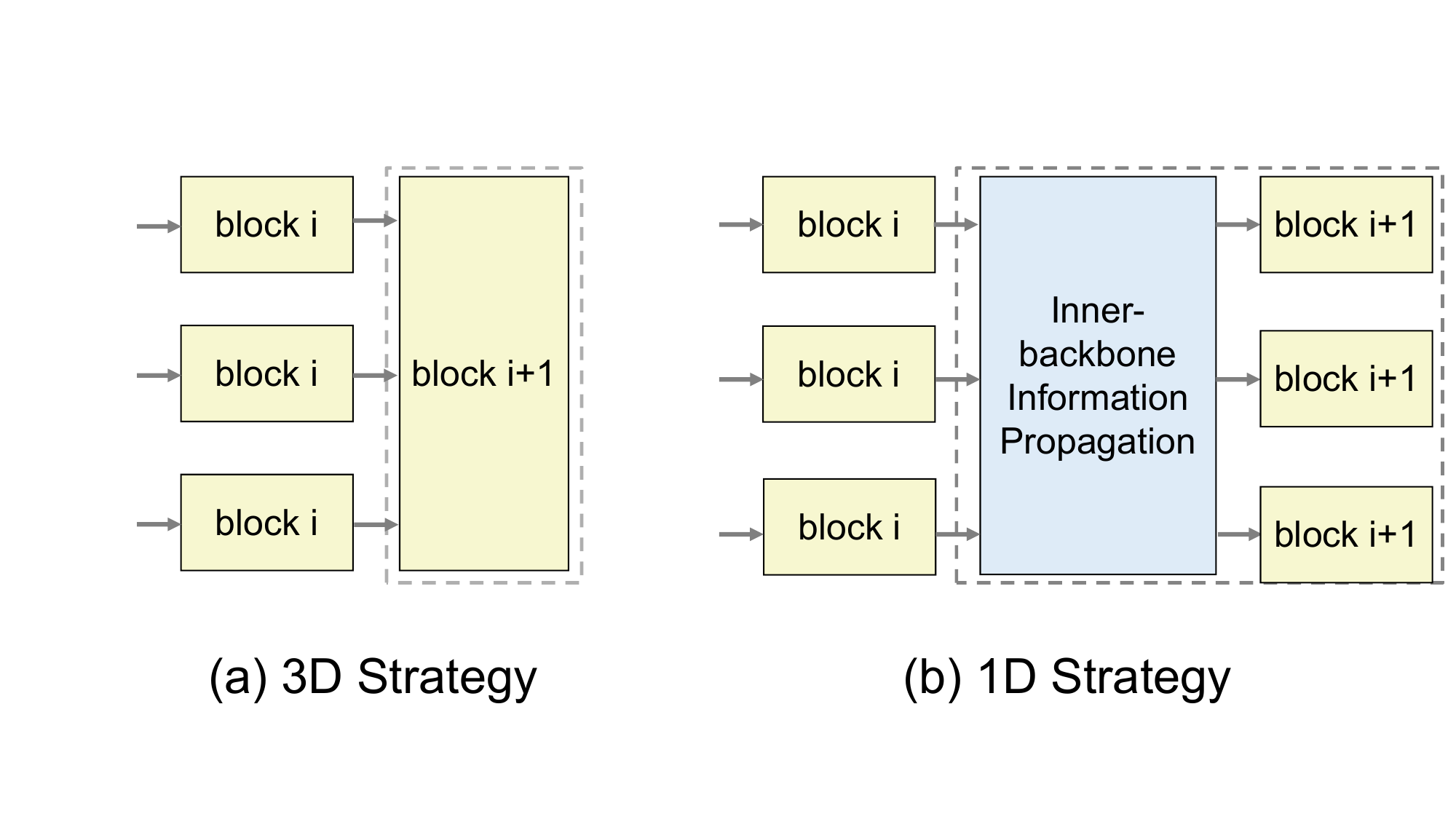}
  \vspace{-5mm}
  \caption{
\textbf{Comparison between 1D and 3D Propagation Strategy.} (a) The 3D setting: the ($i+1$)th block takes all snippets as input and directly applies spatiotemporal self-attention to the whole video clip. (b)The 1D setting: global propagation block is inserted between consecutive backbone blocks.
}
\label{fig:1d_3d}
\end{figure}

\vspace{-4.5mm}
\paragraph{\bf{Study on Global Propagation Strategy.}}
The design of the global block is more flexible than the local block and requires further discussion. As shown in Table~\ref{tbl:positional-encoding}, additional clip-level positional encoding can enable the model to recognize clip-level feature sequences after snippet splicing, leading to better detection results. To save computational consumption, we disassemble the complete cross-snippet spatiotemporal relationship modeling into cross-snippet temporal modeling and inner-snippet spatiotemporal modeling. 
For vivid comparison, our global propagation block is referred to as 1D (temporal only) strategy shown in Fig~\ref{fig:1d_3d}(b), and applying clip-level spatiotemporal modeling on the last block of each subset in the backbone is referred to as 3D (temporal+spatial) strategy (Fig~\ref{fig:1d_3d}(a)). Both 1D and 3D strategies obtain similar detection results while 3D strategy causes more memory consumption shown in Table~\ref{tbl:attn3d}, inferring that concentrating on clip-level temporal modeling is sufficient. We further study where global blocks should be located in the backbone. By default 4 global blocks are placed evenly. We compare by placing them in the first or last 4 blocks instead. As is shown in Table~\ref{tbl:locations-of-propagation-blocks}, evenly 4 blocks configuration performs best, and last 4 blocks configuration follows. This is in line with the observation in ViT~\cite{vit} that ViT has a longer attention distance in later blocks and is more localized in earlier ones. Premature temporal interaction of individual snippet features yields limited benefit. Finally, we adopt 4 evenly placed global blocks as the final configuration of the global propagation block. 

\begin{table}[]
\resizebox{0.47\textwidth}{!}{
\begin{tabular}{c|c|ccccc|c}
\hline
prop. strategy   & use p-b & 0.3  & 0.4  & 0.5  & 0.6  & 0.7  & Avg  \\ 
\hline
\multirow{2}{*}{none} &   &  74.6    & 70.2 &62.6  & 51.3 &38.4   &   59.4 \\ 
   & \checkmark & 77.3 & 73.5 & 65.1 & 52.6 & 37.7 & \textbf{61.2} \\ \hline
\multirow{2}{*}{4 global blocks}    &   & 78.4 & 74.5 & 66.6 & 53.4 & 38.7 & 62.3 \\
 & \checkmark  & 78.7 & 74.1 & 66.2 & 55.5 & 40.3 & \textbf{63.0} \\ \hline
\multirow{2}{*}{12 local blocks}      &  & 77.1 & 72.8 & 65.4 & 54.2 & 40.6 & \textbf{62.0} \\
  & \checkmark & 77.5 & 73.3 & 65.9 & 52.7 & 38.2 & 61.5 \\ \hline
\end{tabular}
}
 \vspace{-3mm}
\caption{\textbf{Study on Post-backbone Information Propagation}. p-b is short for post-backbone information propagation. 
}
\label{table:analysis_ta}
\end{table}

\vspace{-6.0mm}
\paragraph{\bf{Study on Post-backbone Information Propagation.}}
Post-backbone information propagation strategy is used for sufficient feature interaction among snippets. It shows that only applying such a strategy ("none" in Table~\ref{table:analysis_ta}) is also effective. Equipped with post-backbone information propagation, we then compare the detection results for both local and global blocks with the best configuration each in the inner-backbone strategy. 
As is shown in Table~\ref{table:analysis_ta},
we could see that the global propagation strategy obtains better detection results (from 62.3\% to 63.0\%), while the detection results of the local propagation strategy drop (from 62.0\% to 61.5\%), so we employ the global propagation strategy for both inner-backbone and post-backbone information propagation strategies.

\vspace{-6.0mm}
\paragraph{\bf{Comparison between ViT-TAD and other TAD Pipelines.}}

\begin{table}[]
\resizebox{0.47\textwidth}{!}{
\begin{tabular}{c|c|ccccc|c}
\hline
Method       & details           & 0.3  & 0.4  & 0.5  & 0.6  & 0.7  & Avg  \\ \cline{1-8}
\multirow{2}{*}{ActionFormer~\cite{ActionFormer}} & ViT-S$_{224\times224}^{30FPS}$   & 75.2 & 70.0 & 62.0 & 48.7 & 33.4 & 57.9 \\   
             & ViT-S* & 75.6 & 70.3 & 63.4 & 52.4 & 39.6 & \textbf{60.3} \\\hline
\multirow{2}{*}{TriDet~\cite{tridet}} &  ViT-B$_{224\times224}^{30FPS}$  & 82.9 & 78.5& 72.0&61.2 & 46.2&\textbf{68.2}  \\
                        &  ViT-B$_{160\times160}^{8FPS}$   & 77.0 & 72.2& 63.8&51.6 & 35.1&60.0  \\  \hline
    BasicTAD(baseline)~\cite{basictad}         & ViT-S$_{160\times160}^{8FPS}$ & 74.6 & 70.2 & 62.6 & 51.3 & 38.4 & 59.4 \\   
    ViT-TAD         & ViT-S$_{160\times160}^{8FPS}$ & 78.7 & 74.1 & 66.2 & 55.5 & 40.3 & 63.0 \\   
    ViT-TAD   &  ViT-B$_{160\times160}^{8FPS}$  & 85.1 & 80.9 & 74.2 & 61.8 & 45.4 & {\bf 69.5} \\
    \hline
\end{tabular}
}
 \vspace{-3mm}
\caption{\textbf{Comparison between ViT-TAD and other TAD pipelines.} ViT-S* means BasicTAD fine-tunes that backbone feature in an end-to-end manner. Details include the choice of backbone, frame resolution, and frame rate.}
\label{table:analysis_head}
\end{table}

\begin{table}[]
\resizebox{0.47\textwidth}{!}{
\begin{tabular}{c|c|ccccc|c}
\hline
Method     & Backb.  & 0.3 & 0.4 & 0.5 & 0.6 &0.7 & Avg \\ \hline
TALLFormer~\cite{TALLFormer} & Swin-B & 76.0 & -  & 63.2 & - & 34.5 & 59.2  \\
TALLFormer (w.o. proposed module) & ViT-B & 78.9 & 75.0  & 67.6 & 56.1 & 37.9 & 63.1  \\
TALLFormer (w. proposed module) & ViT-B & 81.0 &  77.0  & 70.7 & 58.9 & 42.7 & \textbf{66.1}  \\ \hline
\end{tabular}
}
 \vspace{-3mm}
\caption{\textbf{Effectiveness of proposed module in ViT-TAD to TALLFormer. }
Proposed module means both inner-backbone and post-backbone propagation blocks.
}
\label{table:analysis_e2e_head}
\end{table}

In this section, we compare ViT-TAD with four convincing methods of TAD, namely ActionFormer~\cite{ActionFormer}, BasicTAD~\cite{basictad}, TALLFormer~\cite{TALLFormer} and Tridet~\cite{tridet}.
Here we migrate the feature modeling capabilities of ViT-TAD to these methods to further explore the reasons why ViT-TAD achieves good detection results.
Shown in Table~\ref{table:analysis_head}, we use the feature extracted by ViT-S for ActionFormer, obtaining an average mAP of 57.9\% on THUMOS14. While BasicTAD achieves an average mAP of 59.4\% with a lower frame rate and resolution. To eliminate the benefit of end-to-end training manner, we treat the backbone fine-tuned by BasicTAD as a feature extractor for ActionFormer, making the detection accuracy of ActionFormer reach an average mAP of 60.3\%. When we further adopt our ViT-TAD based on BasicTAD, it can achieve an average mAP of 63.0\%, even higher than ActionFormer with fine-tuned features from BasicTAD. When we further compare our ViT-TAD with the state-of-the-art TAD method TriDet that inputs pre-extracted features, its performance is worse than our ViT-TAD when using pre-extracted ViT-B features based on 30 FPS frame rate and $224\times224$ frame resolution (compare 68.2\% with 69.5\%), and even worse under fair input conditions (compare 60.0\% with 69.5\%). Therefore we can draw several conclusions. First, the Actionformer's detector is stronger than BasicTAD. Second, BasicTAD achieves better results after using the proposed modules in ViT-TAD. Third, ViT-TAD outperforms the state-of-the-art TAD method TriDet under fair input conditions. To further verify the module's effectiveness proposed in ViT-TAD, we migrate it to another end-to-end TAD method TALLFormer that handles clip-level temporal modeling. Shown in Table~\ref{table:analysis_e2e_head}, our proposed modules improve TALLFormer's detection results when both are applied with ViT-B provided by~\cite{videomae_v2} (from 63.1\% to 66.1\%). In summary, ViT-TAD has significant advantages over the above TAD methods, and its performance is better than the current non-end-to-end TAD method TriDet. At the same time, the modules proposed in ViT-TAD can be adopted by other TAD methods to achieve better detection results. 
\vspace{-6.0mm}
\paragraph{\bf{Runtime Comparison with other TAD Methods.}}

\begin{table}[]
\resizebox{0.47\textwidth}{!}{
\begin{tabular}{c|c|ccc}
\hline
Methods&  details & GPU & FPS  & Avg   \\ \hline
ActionFormer~\cite{ActionFormer} &  ViT-B$_{224\times224}^{30FPS}$ & 3090 & 268 & 65.1  \\ 
TriDet~\cite{tridet} &  ViT-B$_{224\times224}^{30FPS}$ & 3090 & 239 & 68.2  \\ \hline
PBRNet~\cite{pbrnet}    &  I3D$_{96\times96}^{10FPS}$ & 1080Ti  &  $<$1488 & 47.1 \\
AFSD~\cite{afsd}    &  I3D$_{96\times96}^{10FPS}$ & 1080Ti  &  $<$3259 & 52.0 \\ 
e2e-tadtr~\cite{e2e-tadtr}    &  res50-SlowFast$_{112\times112}^{10FPS}$ & TITAN Xp  &  5076 & 54.2 \\
BasicTAD~\cite{basictad}    &  SlowOnly$_{112\times112}^{3FPS}$ & TITAN Xp  &  7143 & 54.5 \\
ViT-TAD   &  ViT-S$_{112\times112}^{8FPS}$ & TITAN Xp & 2135 & 63.0 \\ 
ViT-TAD   &  ViT-S$_{160\times160}^{8FPS}$ & TITAN Xp & 1349 & 64.3 \\ 
ViT-TAD   &  ViT-B$_{160\times160}^{8FPS}$ & TITAN Xp & 845 & \textbf{69.5} \\ \hline
\end{tabular}}
\caption{\textbf{Runtime comparison with other TAD methods.}
Details include the choice of backbone, frame resolution, and frame rate. 
}
\label{table:speed}
\vspace{-3mm}
\end{table}

Here we list the runtime comparison with other TAD methods shown in Table~\ref{table:speed}. The time cost of optical flow extraction makes PBRNet~\cite{pbrnet} and AFSD~\cite{afsd} lower than the reported speed. TriDet~\cite{tridet} and ActionFormer~\cite{ActionFormer} need to extract features of the complete video in advance, so we consider the inference speed of both feature extraction and model inference. Since our method is based on ViT, its inference speed cannot be compared with convolution-based methods~\cite{e2e-tadtr,basictad}, but our ViT-TAD achieves the best detection results while ensuring fast inference speed among methods using ViT.

\subsection{Analysis}
\paragraph{\bf{Efficiency Analysis of ViT-TAD.}}
\begin{table}[]
\resizebox{0.47\textwidth}{!}{
\begin{tabular}{cc|ccccc}
\hline
prop. strategy  & details   & Avg  & params & train mem. & FLOPs & FPS \\ \hline
none     &  ViT-S$_{112\times112}^{8FPS}$  & 59.4 &  25.62M      &   8.39G        &   104.59G     &     2204      \\ \hline  
\multirow{3}{*}{4 global blocks+p-b} & ViT-S$_{112\times112}^{8FPS}$ & \textbf{63.0} &  33.35M      &   9.33G        &   108.80G     &     2135 \\
& ViT-S$_{160\times160}^{8FPS}$ & \textbf{64.3} &  33.35M      &   23.56G(4.35G*)        &   220.59G     &     1349 \\
& ViT-B$_{160\times160}^{8FPS}$ & \textbf{69.5} &  131.25M      &   8.97G*       &   866.73G     &     845
\\ \hline  
\end{tabular}
}

 \vspace{-2mm}
\caption{\textbf{Practical performance of backbone adaptation strategies}. p-b is short for post-backbone information propagation. Details include backbone, frame resolution, and frame rate.
$*$ means checkpoint training strategy is used.}
\label{table:analysis}
\end{table}
We compare the differences between baseline and variants of ViT-TAD. 
As is shown in Table~\ref{table:analysis}, 
ViT-TAD introduces additional training parameters and consumes more memory due to the introduction of multiple blocks in the backbone. 
Thanks to checkpoint training strategy (marked by $*$ in Table~\ref{table:analysis}), it can still run successfully with limited computing resources.

\begin{figure}[!t]
  \includegraphics[width=0.45\textwidth]{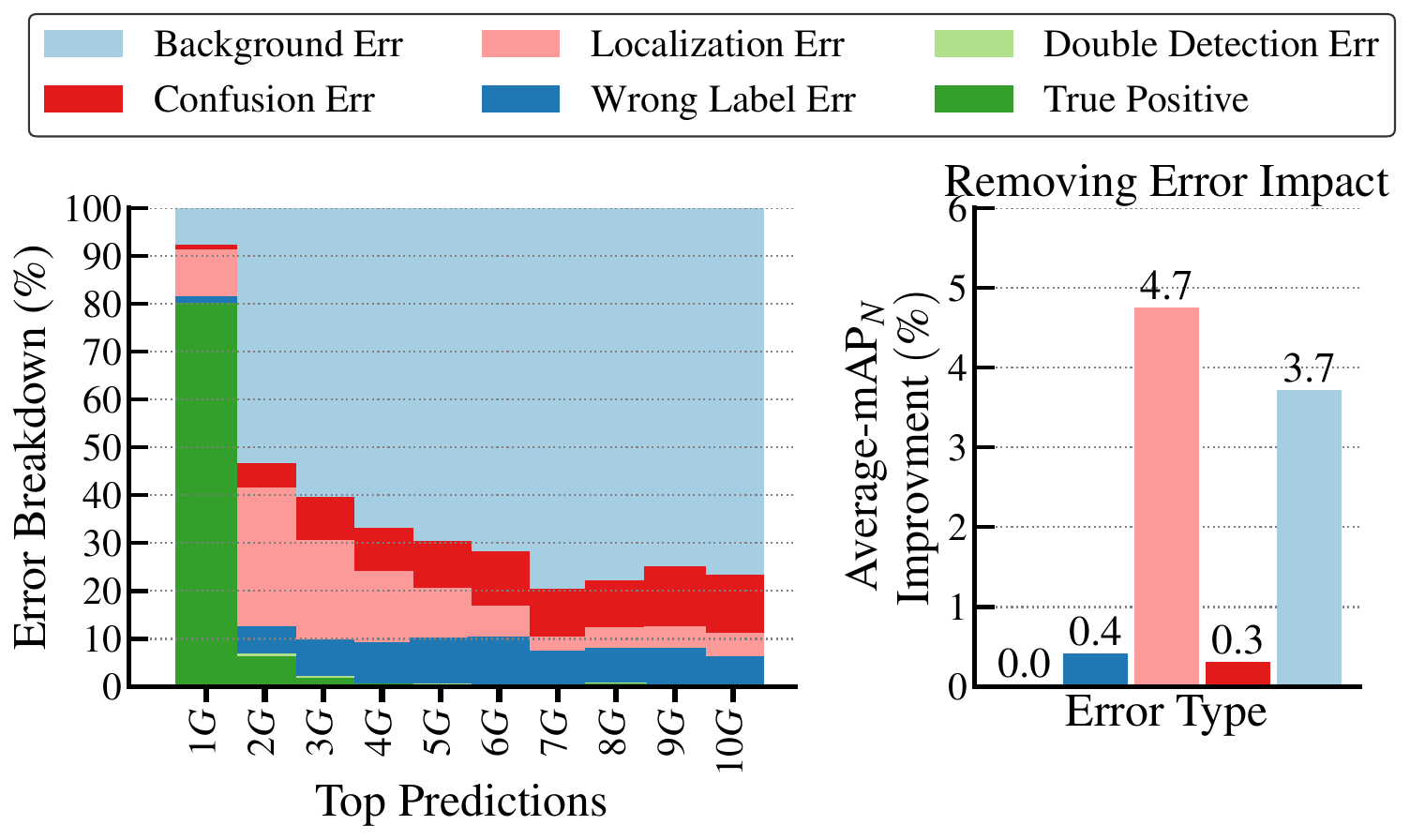}
  \vspace{-4mm}
  \caption{
\textbf{Error analysis of our ViT-TAD.} There are error rates of 5 types on top-10G predictions, where G denotes the number of ground truths.
}
\label{fig:error}
\vspace{-6mm}
\end{figure}

\vspace{-2mm}
\paragraph{\bf{Error Analysis.}}
To analyze the limitations of our model, we provide false positive error chart~\cite{error} of our ViT-TAD on THUMOS14 dataset shown in Fig~\ref{fig:error}. We obtain quite high true positive rate on Top-1G predictions. As is mentioned by BasicTAD~\cite{basictad}, one-stage anchor-free methods suffer from  ``Background Error" due to limited anchors causing more predictions failing to match
ground truth. A more precise regression loss design is needed.

\begin{figure}[!t]
  \includegraphics[width=0.45\textwidth]{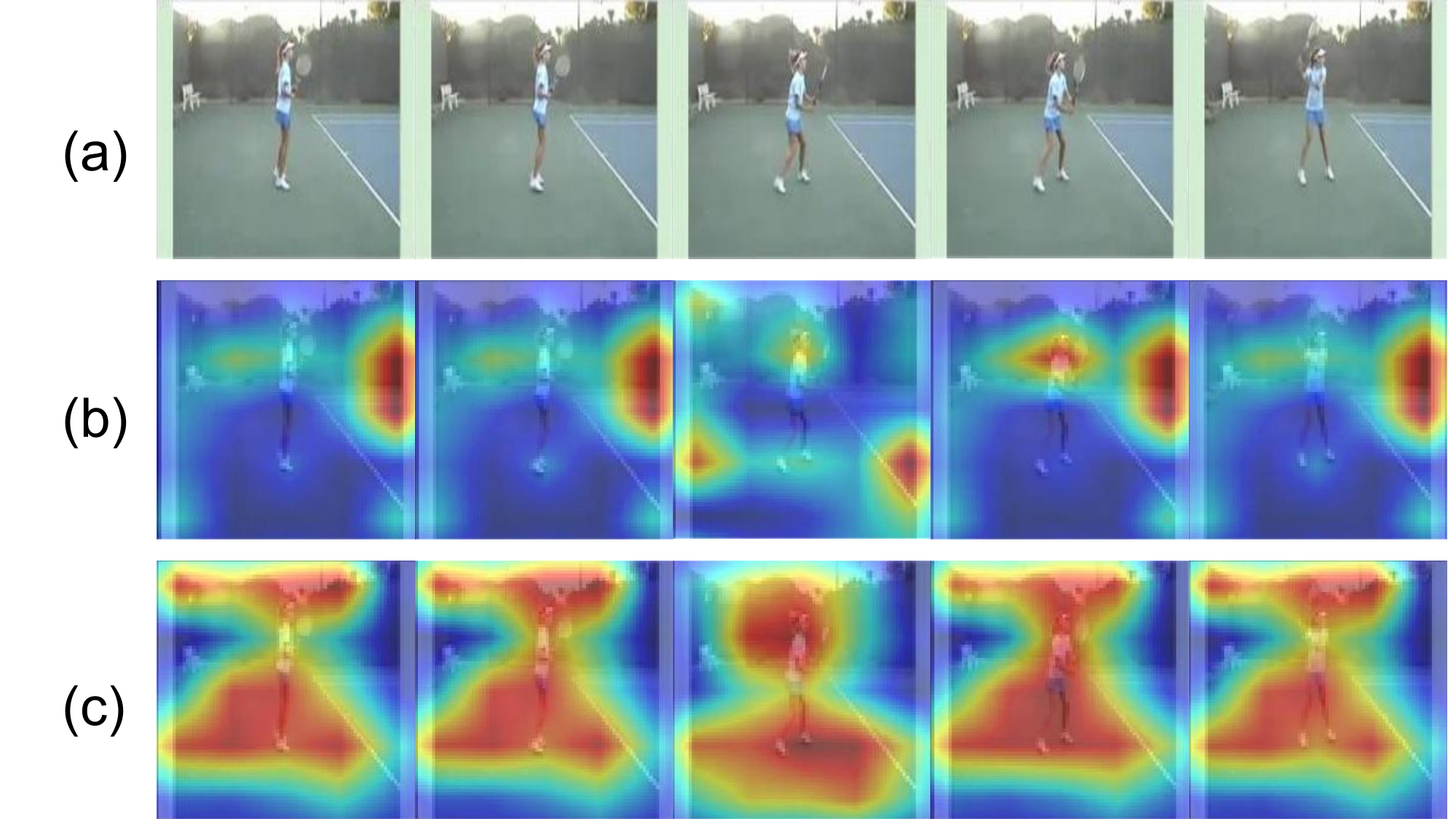}
  \caption{
\textbf{Heatmap visualization.} (a): Original input frames. (b): Heatmap from baseline method. (c): Heatmap from ViT-TAD.
}
\vspace{-3mm}
\label{fig:visual}
\end{figure}

\subsection{Visualization}
We provide heatmap of action ``TennisSwing" in THUMOS14 for both baseline and ViT-TAD detectors to visualize inner-backbone information propagation strategy's impact on frame-level modeling. We use Grad-CAM~\cite{gradcam} to generate corresponding heatmap in backbone's last layer. As is shown in Fig~\ref{fig:visual}, due to the integration of global temporal information, the model prioritizes the regions of the image that are relevant for action recognition, encompassing the athlete's body and the pertinent background during the action. When the model is restricted to short-term observations, it may lead to the model diverting its attention to irrelevant or imprecise areas.

\subsection{Comparison with the State of the Art}

\begin{table}[]
\resizebox{0.48\textwidth}{!}{
\begin{tabular}{c|cc|ccccc|c}
\hline
 Method                               & Backb.    & Flow & 0.3  & 0.4  & 0.5  & 0.6  & 0.7  & Avg  \\ \hline
 BSN~\cite{bsn}                                  & TSN       & \CheckmarkBold     & 53.5 & 45.0 & 36.9 & 28.4 & 20.0 & 36.8 \\
 BMN~\cite{bmn}                                 & TSN       & \CheckmarkBold     & 56.0 & 47.4 & 38.8 & 29.7 & 20.5 & 38.5 \\
  DBG~\cite{dbg}                                  & TSN       & \CheckmarkBold     & 57.8 & 49.4 & 39.8 & 30.2 & 21.7 & 39.8 \\
 BC-GNN~\cite{bcgnn}                                  & TSN       & \CheckmarkBold     & 57.1 & 49.1 & 40.4 & 31.2 & 23.1 & 40.2 \\
 G-TAD~\cite{gtad}                                & TSN       &  \CheckmarkBold    & 66.4 & 60.4 & 51.6 & 37.6 & 22.9 & 47.8 \\
  RTD-Net~\cite{rtd}                              & I3D       & \CheckmarkBold     & 58.5 & 53.1 & 45.1 & 36.4 & 25.0 & 43.6 \\
   VSGN~\cite{vsgn}                                 & TSN       & \CheckmarkBold     & 66.7 & 60.4 & 52.4 & 41.0 & 30.4 & 50.2 \\
    ReAct~\cite{react}                                 & TSN       & \CheckmarkBold     & 69.2 & 65.0 & 57.1 & 47.8 & 35.6 & 55.0 \\
   ActionFormer~\cite{ActionFormer}                        & I3D       & \CheckmarkBold     & 82.1 & 77.8 & 71.0 & 59.4 & 43.9 & 66.8 \\ 
   ActionFormer~\cite{ActionFormer}                        & ViT-B       & \XSolidBrush     &80.5&75.7&68.6&57.9&42.6&65.1\\
TriDet~\cite{tridet}                        & I3D       & \CheckmarkBold     & 83.6 & 80.1 & 72.9 & 62.4 & 47.4 & 69.3 \\ 
VideoMAE V2~\cite{videomae_v2}              & ViT-G     &  \XSolidBrush    & 84.0 & 79.6 & 73.0 & 63.5 & 47.7 & \textbf{69.6} \\
\hline
     
 PBRNet~\cite{pbrnet}$_{96\times96}$                               & I3D       & \CheckmarkBold     & 58.5 & 54.6 & 51.3 & 41.8 & 29.5 & 47.1 \\
 AFSD~\cite{afsd}$_{96\times96}$                                 & I3D       &  \CheckmarkBold    & 67.3 & 62.4 & 55.5 & 43.7 & 31.1 & 52.0 \\
 STPT~\cite{stpt}$_{96\times96}$                              & MViT &  \XSolidBrush    & 70.6 & 65.7 & 56.4 & 44.6 & 30.5 & 53.6 \\
 DaoTAD~\cite{daotad}$_{112\times112}$                              &  res50-I3D &  \XSolidBrush    & 62.8 & 59.5 & 53.8 & 43.6 & 30.1 & 50.0 \\
 e2e-tadtr~\cite{e2e-tadtr}$_{112\times112}$                              &  SlowFast-R50 &  \XSolidBrush    & 69.4 & 64.3 & 56.0 & 46.4 & 34.9 & 54.2 \\
 TALLFormer~\cite{TALLFormer}$_{224\times224}$                   &  Swin-B     &  \XSolidBrush    & 76.0 & -    & 63.2 & -    & 34.5 & 59.2 \\
 TALLFormer~\cite{TALLFormer}$_{224\times224}$                   &  ViT-B     &  \XSolidBrush    & 78.9 & 75.0    & 67.6 & 56.1    & 37.9 & 63.1 \\
 BasicTAD~\cite{basictad}$_{160\times160}$                    & SlowOnly-R50     &  \XSolidBrush    & 75.5 & 70.8    & 63.5 & 50.9    & 37.4 & 59.6 \\
Re$^{2}$TAL~\cite{re2tal}$_{224\times224}$ & Swin-T     &  \XSolidBrush    & 77.0 & 71.5 & 62.4 & 49.7 & 36.3 & 59.4 \\
 Re$^{2}$TAL~\cite{re2tal}$_{224\times224}$&  SlowFast-R101  &  \XSolidBrush    & 77.4 & 72.6 & 64.9 & 53.7 & 39.0 & 61.5 \\
 BasicTAD~\cite{basictad}(baseline)$_{112\times112}$ & ViT-S & \XSolidBrush & 74.6 &70.2 &62.6 & 51.3 & 38.4 & 59.4 \\
 ViT-TAD$_{112\times112}$                              & ViT-S     &  \XSolidBrush    & 78.7 & 74.1 & 66.2 & 55.5 & 40.3 & 63.0 \\
 ViT-TAD$_{160\times160}$                              & ViT-S     &  \XSolidBrush    & 79.8 & 75.2 & 68.4 & 56.4 & 41.7 & 64.3 \\ 
 ViT-TAD$_{160\times160}$                              & ViT-B     &  \XSolidBrush    & 85.1 & 80.9 & 74.2 & 61.8 & 45.4 & {\bf 69.5} \\ \hline
\end{tabular}
}
\vspace{-3mm}
\caption{\textbf{Comparison with state-of-the-art methods on THUMOS14.} The subscript indicates the spatial resolution. \textbf{Flow} denotes whether each method uses optical flow as input.
}
\label{tbl:sota-thumos}
\end{table}

\begin{table}[]

\resizebox{0.48\textwidth}{!}{
\begin{tabular}{c|cc|ccc|c}
\hline
Method                                                 & Backb.  & Flow                          & 0.5   & 0.75  & 0.95 & Avg   \\ \hline
BSN~\cite{bsn}                                                    & TSN     & \CheckmarkBold & 46.45 & 29.96 & 8.02 & 30.03 \\
ReAct~\cite{react}                                                    & TSN     & \CheckmarkBold & 49.60 & 33.00 & 8.60 & 32.60 \\
BMN~\cite{bmn}                                                    & TSN     & \CheckmarkBold & 50.07 & 34.78 & 8.29 & 33.85 \\
BC-GNN~\cite{bcgnn}                                                    & TSN     & \CheckmarkBold & 50.56 & 34.75 & 9.37 & 34.26 \\
G-TAD~\cite{gtad}                                                  & TSN     & \CheckmarkBold & 50.36 & 34.60 & 9.02 & 34.09 \\
RTD-Net~\cite{rtd}                                                & I3D     & \CheckmarkBold & 47.21 & 30.68 & 8.61 & 30.83 \\
VSGN~\cite{vsgn}                                                   & TSN     & \CheckmarkBold & 52.38 & 36.01 & 8.37 & 35.07 \\
ActionFormer~\cite{ActionFormer}                                           & I3D     & \CheckmarkBold & 53.50 & 36.20 & 8.20 & 35.60 \\ 
TriDet~\cite{tridet}                                           & I3D     & \CheckmarkBold & 54.50 & 36.80 & 11.50 & 36.80 \\ 
TriDet~\cite{tridet}                                           & SlowFast     & \XSolidBrush & 56.70 & 39.30 & 11.70 & \textbf{38.60} \\ \hline
STPT~\cite{stpt}$_{96\times96}$                                             & MViT & \XSolidBrush   & 51.40 & 33.70 & 6.80 & 33.40 \\
AFSD~\cite{afsd}$_{96\times96}$                                                   & I3D     & \CheckmarkBold & 52.40 & 35.30 & 6.50 & 34.40 \\
PBRNet~\cite{pbrnet}$_{96\times96}$                                                 & I3D     & \CheckmarkBold & 53.96 & 34.97 & 8.98 & 35.01 \\
e2e-tadtr~\cite{e2e-tadtr}$_{112\times112}$                                             & SlowFast-R50 & \XSolidBrush   & 50.47 & 35.99 & 10.83 & 35.10 \\
TALLFormer~\cite{TALLFormer}$_{224\times224}$                                             & Swin-B & \XSolidBrush   & 54.10 & 36.20 & 7.90 & 35.60 \\
Re$^{2}$TAL~\cite{re2tal}$_{224\times224}$  & Swin-T & \XSolidBrush   & 54.75 & 37.81 & 9.03 & 36.80 \\
Re$^{2}$TAL~\cite{re2tal}$_{224\times224}$  & SlowFast-R101 & \XSolidBrush   & 55.25 & 37.86 & 9.05 & 37.01 \\
ViT-TAD$_{160\times160}$                                                 & ViT-S   & \XSolidBrush   & 55.09 & 37.81 & 8.75 & 36.69 \\ 
ViT-TAD$_{160\times160}$                                                 & ViT-B   & \XSolidBrush   & 55.87 & 38.47 & 8.80 & \textbf{37.40} \\ \hline
\end{tabular}
}
\vspace{-3mm}
\caption{\textbf{Comparison with state-of-the-art methods on ActivityNet-1.3.} The subscript indicates the spatial resolution. \textbf{Flow} denotes whether each method uses optical flow as input.
}
\label{table:sota-anet}
\end{table}
\begin{table}[]
\resizebox{0.48\textwidth}{!}{
\begin{tabular}{c|cc|ccc|c}
\hline
Method   & Backbone     & Flow & 0.5   & 0.75  & 0.95 & Avg   \\ \hline
BMN~\cite{bmn}      & I3D          &  \CheckmarkBold    & 14.44 & 8.92  & 3.12 & 9.25  \\
DBG~\cite{dbg}      & I3D          &  \CheckmarkBold    & 10.65 & 6.43  & 2.50 & 6.75  \\
G-TAD~\cite{gtad}    & I3D          &  \CheckmarkBold    & 13.74 & 8.83  & 3.06 & 9.06  \\
ActionFormer~\cite{ActionFormer} & I3D &  \XSolidBrush    & - & - & - & 13.20 \\
VideoMAE V2~\cite{videomae_v2} & ViT-G &  \XSolidBrush    & 29.07 & 17.66 & 5.07 & \textbf{18.24} \\
\hline
BasicTAD~\cite{basictad}$_{160\times160}$ & SlowOnly-R50 &  \XSolidBrush    & 24.34 & 10.57 & 0.43 & 12.15 \\ 
ViT-TAD$_{160\times160}$  & ViT-B        &  \XSolidBrush    & 32.61     &  15.85     &  2.68    &  \textbf{17.20}     \\ \hline
\end{tabular}
}
\vspace{-3mm}
\caption{\textbf{Comparison with state-of-the-art methods on FineAction.} The subscript indicates the spatial resolution. \textbf{Flow} denotes whether each method uses optical flow as input.
}
\vspace{-3mm}
\label{tbl:sota-fineaction}
\end{table}

We compare our ViT-TAD with the previous state-of-the-art methods on THUMOS14~\cite{thumos14}, ActivityNet-1.3~\cite{anet} and FineAction~\cite{fineaction}. We classify these methods into end-to-end methods (lower part of tables) and non-end-to-end methods (upper part of tables). For THUMOS14, the results are shown in Table~\ref{tbl:sota-thumos}. Our ViT-TAD with ViT-B outperforms all TAD methods except VideoMAE V2 which adopts ViT-G. 
For ActivityNet-1.3, the results are shown in Table~\ref{table:sota-anet}. 
We attempt to adapt our model to predict binary action proposals and obtain the detection results by applying video-level action classifiers~\cite{cuhk}. We get competitive results with ViT-S (36.69\%) and state-of-the-art results with ViT-B (37.40\%) among all end-to-end TAD methods. 
For FineAction, the results are shown in Table~\ref{tbl:sota-fineaction}. We get good results with ViT-B (17.20\%), only slightly lower than VideoMAE V2~\cite{videomae_v2}.

\section{Conclusion}
In this paper, we have presented a simple TAD framework (ViT-TAD) based on the plain ViT backbone. Our ViT-TAD incorporates the inner-backbone propagation module and post-backbone propagation module to capture more fine-grained temporal information across different snippets and global contexts. We perform in-depth ablation studies on the design of different components in ViT-TAD.
With the simple TAD head and powerful masked video pre-training, our ViT-TAD yields a state-of-the-art performance compared with other end-to-end methods on the challenging datasets THUMOS14, ActivityNet-1.3 and FineAction. We hope it will serve as a new TAD baseline for future research.

\noindent {\small  \textbf{Acknowledgements.}  This work is supported by the National Key R$\&$D Program of China (No. 2022ZD0160900), the National Natural Science Foundation of China (No. 62076119, No. 61921006), and Collaborative Innovation Center of Novel Software Technology and Industrialization.}

{\small
\bibliographystyle{ieee_fullname}
\bibliography{egbib}
}

\end{document}